\documentclass[a4paper,fleqn]{cas-dc}
\usepackage[utf8]{inputenc}
\usepackage[T1]{fontenc}
\usepackage{amsmath}
\usepackage{amssymb}
\usepackage{graphicx}
\usepackage{url}
\usepackage{booktabs}
\usepackage{multirow}
\usepackage{float}
\usepackage[caption=false]{subfig}
\usepackage{amssymb}
\usepackage{hyperref}
\usepackage[capitalize]{cleveref}
\crefname{section}{Sec.}{Secs.}
\Crefname{section}{Section}{Sections}
\Crefname{table}{Table}{Tables}
\crefname{table}{Tab.}{Tabs.}
\crefname{algocf}{Algo.}{Algos.}
\Crefname{algocf}{Algorithm}{Algorithms}

\usepackage{paralist}
\usepackage[numbers]{natbib}
\usepackage{threeparttable}
\usepackage[vlined ,ruled,linesnumbered]{algorithm2e}

\definecolor{myblue}{RGB}{100, 140, 200}
\newcommand{\cb}{}
\usepackage{booktabs,amsfonts,dcolumn}

\SetKwFor{Loop}{Loop}{}{EndLoop} %

\def\tsc#1{\csdef{#1}{\textsc{\lowercase{#1}}\xspace}}
\tsc{WGM}
\tsc{QE}
\tsc{EP}
\tsc{PMS}
\tsc{BEC}
\tsc{DE}

\begin{document}
\let\WriteBookmarks\relax
\def\floatpagepagefraction{1}
\def\textpagefraction{.001}

\shorttitle{DyCE: Dynamically Configurable Exiting for Deep Learning Compression and Real-time Scaling}

\shortauthors{Qingyuan Wang et~al.}

\title [mode = title]{DyCE: Dynamically Configurable Exiting\\ for Deep Learning Compression and Real-time Scaling
}

\tnotemark[1,2]

\tnotetext[1]{This publication has emanated from research supported in part by 1) a grant from Science Foundation Ireland under Grant number 18/CRT/6183 and 2) CHIST-ERA grant JEDAI CHIST-ERA-18-ACAI-003. For the purpose of Open Access, the author has
    applied a CC BY public copyright licence to any Author Accepted Manuscript version arising from
    this submission.}
\tnotetext[2]{Code available at \url{https://github.com/QingyuanWang/dyce}}

\author[1]{Qingyuan Wang}[type=editor,
    orcid=0000-0002-7879-4328]

\cormark[1]
\ead{qingyuan.wang@ucdconnect.ie}

\author[1]{Barry Cardiff}[type=editor,
    orcid=0000-0003-1303-8115]
\ead{barry.cardiff@ucd.ie}
\author[2]{Antoine Frappé}[type=editor,
    orcid=0000-0002-0977-549X]
\ead{antoine.frappe@junia.com}
\author[2]{Benoit Larras}[type=editor,
    orcid=0000-0003-2501-8656]
\ead{benoit.larras@junia.com}
\author[1]{Deepu John}[type=editor,
    orcid=0000-0002-6139-1100]
\ead{deepu.john@ucd.ie}

\affiliation[1]{organization={University College Dublin},
    city={Dublin},
    country={Ireland}}
\affiliation[2]{organization={Univ. Lille, CNRS, Centrale Lille, Junia, Univ. Polytechnique Hauts-de-France, UMR 8520-IEMN},
    city={Lille},
    country={France}}

\cortext[cor1]{Corresponding author}

\begin{abstract}
    Conventional deep learning (DL) model compression {\cb methods affect all input samples equally. However, as samples vary in difficulty, a dynamic model that adapts computation based on sample complexity offers a novel perspective for compression and scaling. Despite this potential, existing dynamic techniques are typically monolithic and have model-specific implementations, limiting their generalizability as broad compression and scaling methods. Additionally, most deployed DL systems are fixed, and unable to adjust once deployed.
    This paper introduces DyCE, a dynamically configurable system that can adjust the performance-complexity trade-off of a DL model at runtime without needing re-initialisation or re-deployment. DyCE achieves this by adding exit networks to intermediate layers, thus allowing early termination if results are acceptable. DyCE also decouples the design of exit networks from the base model itself, enabling its easy adaptation to new base models. We also propose methods for generating optimized configurations and determining exit network types and positions for desired trade-offs. By enabling simple configuration switching, DyCE provides fine-grained performance tuning in real-time.
    We demonstrate the effectiveness of DyCE through image classification tasks using deep convolutional neural networks (CNNs). DyCE significantly reduces computational complexity by 26.2\% for ResNet$_{\mathit{152}}$, 26.6\% for ConvNextv2$_{\mathit{tiny}}$ and 32.0\% for DaViT$_{\mathit{base}}$ on ImageNet validation set, with accuracy reductions of less than 0.5\%.}
\end{abstract}


\begin{highlights}
    \item {\cb Effectively compress the computational complexity of deep learning models.}
    \item Enable early exiting on any existing deep learning models by attaching tiny exits.
    \item Dynamically change the exiting configuration to achieve arbitrary scaling and adapt the varying demand in real-time.
    \item Generating the best exiting configuration of a multi-exit system for various trade-off preferences.
\end{highlights}

\maketitle
\section{Introduction}
Deep learning (DL) models have exhibited success across a multitude of applications. Nonetheless, they require substantial computations within a short timeframe, which presents significant challenges for deployment on resource-constrained devices, such as Internet of Things (IoT) devices and smartphones. These devices, with their limited computing and memory resources, struggle to accommodate the deployment of large DL models. In addition, the run-time latencies and power consumption would be unacceptable when large DL models are deployed in such environments. Even resource-rich systems, such as data centers, contend with issues regarding energy consumption\cite{dayarathna_data_2016}. Therefore, optimizing model complexity is a critical prerequisite for implementing DL techniques on practical systems.

Various model compression and optimization strategies, like quantization \cite{wu_mixed_2018} and pruning \cite{blalock_what_2020}\cite{lin_filter_2021}, have been introduced to address this complexity concern. Despite their success to varying degrees, they overlook the impact of the difficulty of input data while optimizing model complexity. In practice, input data can range in difficulty levels, some instances being more challenging than others. Difficult input samples might necessitate deep models for acceptable outcomes, while easier inputs might suffice with smaller (shallower) DL models. However, most existing applications use a static data path for all inputs, thereby wasting computational resources on simpler inputs. Routing input samples to various sub-models based on their difficulty level is better suited for allocating computational resources efficiently. Early exit \cite{teerapittayanon_branchynet_2016, kaya_shallow-deep_2019} is one of the most promising dynamic techniques due to its simplicity and flexibility. Typically, several auxiliary exiting networks are connected to the base models, generating outputs from intermediate points in the base model. The computation ceases at the first exit, where the output satisfies an acceptability criterion.
However, most early exit based dynamic models are monolithic designs. i.e. Their components, such as backbone models, exit branches, and exit policies, are tightly coupled. This coupling makes it challenging to apply these designs to other or future models. Additionally, these designs often sacrifice some performance to incorporate dynamic features, limiting their effectiveness as compression methods.

\begin{figure*}[ht]
    \centering
    \includegraphics[width=\linewidth]{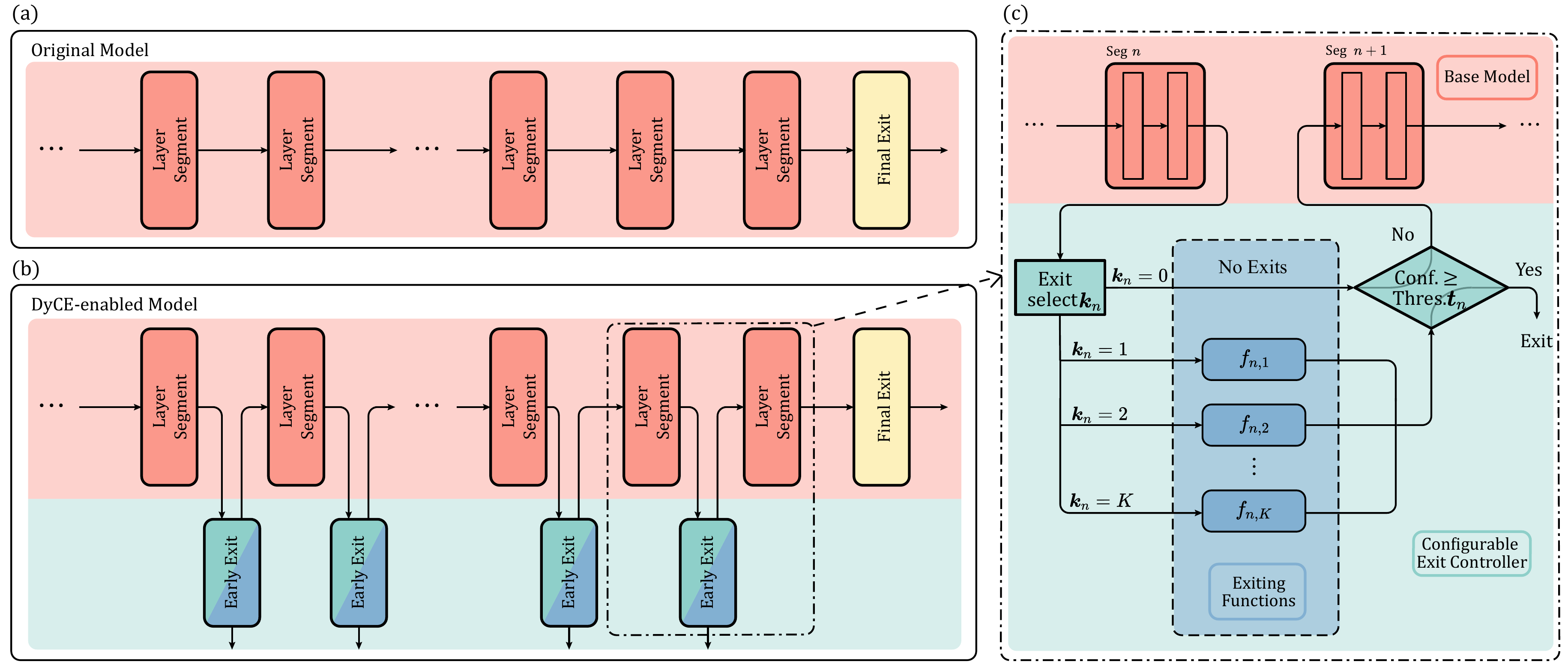}
    \caption{An overview of DyCE system. (a) is the original deep learning model. (b) is the model when DyCE is applied. The backbone of the original model (shown in red) is divided into segments, and multiple early exits (shown in blue) are attached to the end of each segment. The logic of exiting controllers is illustrated in (c). If the confidence level of an output from the exit point is greater than its associated threshold, the inference will complete immediately. Otherwise, the output of the previous backbone segment will be passed to the next segment. The exit select ($k_n$) and corresponding thresholds ($t_n$) are from the selected configuration. These configurations are pre-defined to adapt varying performance-complexity targets.
    }
    \label{fig:early-exit-system}
\end{figure*}

Furthermore, most DL systems, particularly those with static models, encounter a common issue: once designed and deployed in a system, their compression ratio and performance are fixed. Adjustments require redeploying another model. In real-world applications, performance requirements can vary based on environmental or operational conditions. This issue persists for many dynamic models \cite{chen_learning_2020,dai_epnet_2020,wang_skipnet_2018,lin_runtime_2017,fedusSwitchTransformersScaling2022}, as their routing strategies are trained. Therefore, they are not configurable post-training or in real-time. Although some other dynamic models \cite{teerapittayanon_branchynet_2016,hong_panda_2021,hang_msnet_2023,wangNotAllImages2021,chenCFViTGeneralCoarsetoFine2022} allow configuring their routing strategies externally, they typically use a single threshold across all exits to modify the performance-complexity trade-off. Given that different exits are relatively independent and have suitable working conditions individually, this uniform approach is rigid and inefficient for systems with multiple exits. A more flexible routing strategy is needed to achieve optimal real-time scaling.

This work introduces DyCE, a \underline{Dy}namically \underline{C}onfigurable \underline{E}xiting framework. DyCE is an early-exit DL framework that can dynamically configure the performance-complexity trade-off point of a DL model during run-time, without re-initialising or re-deploying a updated model on an inference hardware.
DyCE enhances existing base models by adding arbitrary exit points at any position, eliminating the necessity for re-training or redesigning the original model. Unlike other efficient designs that create new dynamic models, DyCE can transform static models into dynamic ones. This capability allows the latest advanced deep learning models, which are typically static, to benefit from dynamic data path. The flexibility of DyCE in both the system building and running phases makes it a versatile method for model compression and scaling.
Designing an early exit system necessitates answering three challenging questions: (1) How to construct efficient exit networks? (2) Where should the exits be positioned? and (3) When should the system exit? However, solutions to any one of these questions could affect the others, thereby making the design process complex. Furthermore, most existing designs are dedicated to specific base models, often requiring a complete system redesign for the adaptation to any new base models.

The search algorithms in the proposed framework will identify the optimal combination of exits and exit conditions for a given performance-complexity target. Thus, questions (2) and (3) are disassociated from question (1), enabling designers to focus on single exit performance without any potential adverse effects on the overall system. \cref{fig:early-exit-system} provides an overview of the proposed system. The DyCE system can be outlined in three components.
\begin{enumerate}
    \item Early exit functions (labelled in blue): These are represented by various functions attached to the hidden layers of the original model, facilitating an early estimate of the final result. If the output produced at an exit is deemed satisfactory, the inference process concludes at that early layer.
    \item A base DL model (labelled in red): DyCE does not necessitate re-training or fine-tuning of the base model but only requires the base model to be divisible along its depth, which is the case commonly found in most DL models. Thus, the proposed compression system has the significant advantage of being independent of the original model in practical implementations.
    \item The exit controller (labelled in green): This is a major difference between DyCE and other early exit systems. The $n^{\text{th}}$ exit is controlled by $k_n$ and $t_n$, means that exit function $f_{n,k_n}$ is used with the threshold $t_n$. It features a configurable exit controller wherein the exit selection and acceptance criteria are controlled by pre-configured configurations generated via search algorithms to meet any performance-complexity targets.
\end{enumerate}

\begin{figure}[ht]
    \centering
    \includegraphics[width=\linewidth]{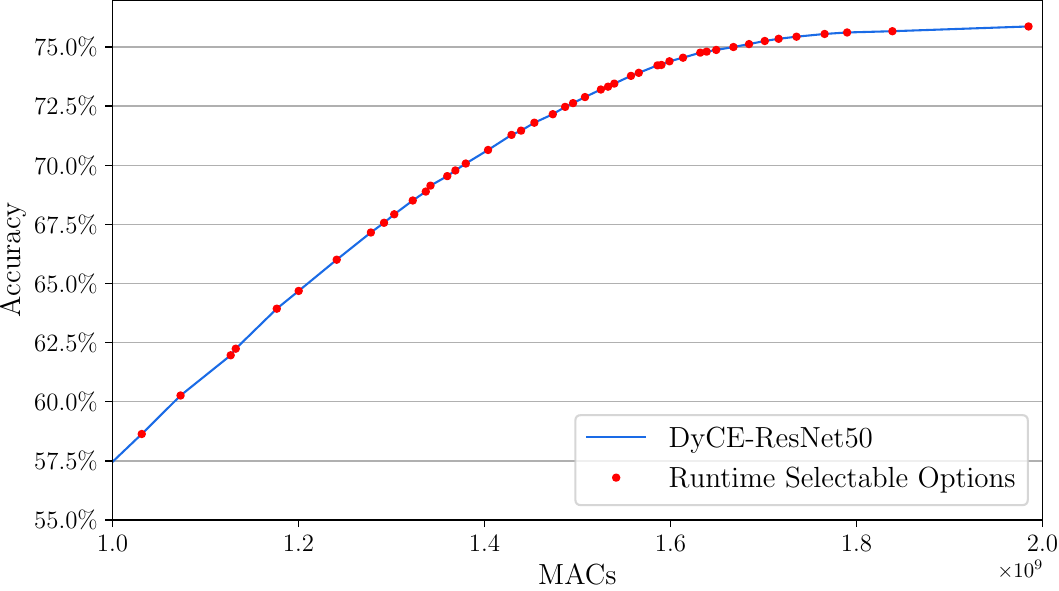}
    \caption{An example of DyCE with ResNet$_{50}$. All red dots selectable performance-complexity operating points in real-time.
    }
    \label{fig:selectable-example}
\end{figure}

In the proposed system, we transform a static model (\cref{fig:early-exit-system} (a)) into a dynamic model (\cref{fig:early-exit-system} (b)) by adding additional networks at numerous potential positions while maintaining the integrity of the original model. {\cb As shown in (\cref{fig:early-exit-system} (c)), the exiting behavior is controlled by the exit function index $k_n$ and the threshold $t_n$ for each position. Modifying these variables changes the system's performance. The specific values for each $k_n$ and $t_n$ are searched configurations. The search algorithm discussed in \cref{sec:config-search} provides configurations for different preferences in fine-grain. During the runtime, the system loads exits and apply the thresholds according to $k_n$ and $t_n$ of a specific configuration. The dynamic scalability is achieved by switching between different configurations.} \cref{fig:selectable-example} is an example of possible configurations (marked as red dots) that DyCE creates for a base model. The density of options can be further increased by using a smaller searching step while creating configurations.
In real-world scenarios, performance requirements can vary over time. Using static compression methods that generate fixed model variants necessitates the replacement of entire models when performance targets shift. This requires hardware devices to store multiple compressed variants of the same model and spend significant resources to re-load different models with varying compression rates into memory. In contrast, the DyCE system can be effortlessly reconfigured on-the-fly by simply switching configurations, allowing for real-time adaptation of model compression rate. This enables the dynamic selection of model performance and complexity trade-off points, gives broad flexibility to the DL model, and ensures the full restoration of the original performance when necessary.

The main contributions of this paper are summarized as follows:
{\cb
\begin{enumerate}
    \item \textbf{DyCE Framework}: The introduced DyCE framework can be applied and dynamically scale and compress existing DL models.
    \item \textbf{Computational Saving}: DyCE can reduce the computation by 26.2\% for ResNet$_{\mathit{152}}$, 26.6\% for ConvNextv2$_{\mathit{tiny}}$ and 32.0\% for DaViT$_{\mathit{base}}$ on ImageNet, with accuracy reductions of less than 0.5\%.
    \item \textbf{Real-time Scaling}: Supporting the selection of complexity-accuracy points to adapt to real-time demands flexibly.
    \item \textbf{Decoupled Design}: DyCE decouples the design considerations for creating an early-exit system, simplifying the developing process.
    \item \textbf{Search Algorithms}: Search algorithms are also proposed to generate configurations for different tradeoff preferences.
\end{enumerate}
}

For evaluating compression performance, we present the results from an image classification task using ImageNet\cite{russakovsky_imagenet_2015}. The results indicate that DyCE, even without a customised exit network design, can adequately harness the potential of attached exits to achieve competitive performance.

The rest of this paper is structured as follows: \cref{sec:related-work} reviews the related compression and scaling methods. \cref{sec:system-description} discusses the inference process proposed in DyCE. \cref{sec:design} explores considerations in the design of exit networks. \cref{sec:config-search} introduces the search algorithm proposed for generating configurations. Finally, the experimental results are discussed in \cref{sec:simulation}.

\section{Related Work}
\label{sec:related-work}

\begin{table*}[htbp]

    \caption{Feature comparison between the proposed method and other compression or scaling methods}
    \label{tab:related-work}
    \resizebox{\linewidth}{!}{
        \begin{threeparttable}[b]
            \begin{tabular}{|c|cc|c|c|c|c|c|c|}
                \hline
                Category                                                                               & \multicolumn{1}{c|}{Method}                                                                                  & Work                                                                                                                                    & \begin{tabular}[c]{@{}c@{}}Scale\\      Up/Down\end{tabular} & \begin{tabular}[c]{@{}c@{}}Loosely\\      Coupled\tnote{1}\end{tabular} & \begin{tabular}[c]{@{}c@{}}No Required \\Backbone Training \end{tabular} & \begin{tabular}[c]{@{}c@{}}Post-training\\      Configurable\tnote{1}\end{tabular} & \begin{tabular}[c]{@{}c@{}}Real-time\\      Configurable\tnote{1}\end{tabular} & \begin{tabular}[c]{@{}c@{}}Fine-grained\\      Tuning\tnote{1}\end{tabular} \\ \hline
                \multirow{8}{*}{\begin{tabular}[c]{@{}c@{}}Utilizing\\ dynamic\\ models \end{tabular}} & \multicolumn{1}{c|}{\multirow{3}{*}{\begin{tabular}[c]{@{}c@{}}Confidence-based \\ Early Exit\end{tabular}}} & \textbf{DyCE}                                                                                                                           & Down                                                         & \textcolor[RGB]{65,174,60}{\checkmark}                                  & \textcolor[RGB]{65,174,60}{\checkmark}                                   & \textcolor[RGB]{65,174,60}{\checkmark}                                             & \textcolor[RGB]{65,174,60}{\checkmark}                                         & \textcolor[RGB]{65,174,60}{\checkmark}                                      \\ \cline{3-9}
                                                                                                       & \multicolumn{1}{c|}{}                                                                                        & \begin{tabular}[c]{@{}c@{}}BranchyNet\cite{teerapittayanon_branchynet_2016} MSDNet\cite{hong_panda_2021} \\ MSNet\cite{hang_msnet_2023}
                                                                                                                                                                                                                            DVT\cite{wangNotAllImages2021} CF-ViT\cite{chenCFViTGeneralCoarsetoFine2022}\end{tabular} & Down                                                         & \textcolor[RGB]{237,90,101}{×}                                          & \textcolor[RGB]{237,90,101}{×}                                           & \Large{-}\tnote{\tiny{2}}                                                          & \Large{-}\tnote{\tiny{2}}                                                      & \Large{-}\tnote{\tiny{2}}                                                                                                 \\ \cline{2-9}
                                                                                                       & \multicolumn{1}{c|}{\begin{tabular}[c]{@{}c@{}}Network-controlled\\ Early Exit\end{tabular}}                 & Chen et al.\cite{chen_learning_2020} EPNet\cite{dai_epnet_2020}                                                                         & Down                                                         & \textcolor[RGB]{237,90,101}{×}                                          & \textcolor[RGB]{237,90,101}{×}                                           & \textcolor[RGB]{237,90,101}{×}                                                     & \textcolor[RGB]{237,90,101}{×}                                                 & \textcolor[RGB]{237,90,101}{×}                                              \\ \cline{2-9}
                                                                                                       & \multicolumn{1}{c|}{Layer Skipping}                                                                          & SkipNet\cite{wang_skipnet_2018}                                                                                                         & Down                                                         & \textcolor[RGB]{237,90,103}{×}                                          & \textcolor[RGB]{237,90,101}{×}                                           & \textcolor[RGB]{237,90,101}{×}                                                     & \textcolor[RGB]{237,90,101}{×}                                                 & \textcolor[RGB]{237,90,101}{×}                                              \\ \cline{2-9}
                                                                                                       & \multicolumn{1}{c|}{Channel Skipping}                                                                        & RNP\cite{lin_runtime_2017}                                                                                                              & Down                                                         & \textcolor[RGB]{237,90,104}{×}                                          & \textcolor[RGB]{237,90,101}{×}                                           & \textcolor[RGB]{237,90,101}{×}                                                     & \textcolor[RGB]{237,90,101}{×}                                                 & \textcolor[RGB]{237,90,101}{×}                                              \\\cline{2-9}
                                                                                                       & \multicolumn{1}{c|}{Sparse MoE\cite{fedusSwitchTransformersScaling2022}}                                     & V-MoE\cite{riquelmeScalingVisionSparse2021}Mobile V-MoE\cite{daxbergerMobileVMoEsScaling2023}                                           & Both\tnote{\tiny{3}}                                         & \textcolor[RGB]{237,90,104}{×}                                          & \textcolor[RGB]{237,90,101}{×}                                           & \textcolor[RGB]{237,90,101}{×}                                                     & \textcolor[RGB]{237,90,101}{×}                                                 & \textcolor[RGB]{237,90,101}{×}                                              \\ \hline
                \multirow{4}{*}{General}                                                               & \multicolumn{2}{c|}{Architecture   Scaling\cite{he_deep_2016}\cite{tan_efficientnet_2019}}                   & Both                                                                                                                                    & \textcolor[RGB]{237,90,105}{×}                               & \textcolor[RGB]{237,90,101}{×}                                          & \textcolor[RGB]{237,90,101}{×}                                           & \textcolor[RGB]{237,90,101}{×}                                                     & \textcolor[RGB]{65,174,61}{\checkmark}                                                                                                                       \\ \cline{2-9}
                                                                                                       & \multicolumn{2}{c|}{Knowledge distillation\cite{hinton_distilling_2015}}                                     & Down                                                                                                                                    & \textcolor[RGB]{65,174,60}{\checkmark}                       & \textcolor[RGB]{237,90,101}{×}                                          & \textcolor[RGB]{237,90,101}{×}                                           & \textcolor[RGB]{237,90,101}{×}                                                     & \textcolor[RGB]{65,174,61}{\checkmark}                                                                                                                       \\   \cline{2-9}                       & \multicolumn{2}{c|}{Pruning\cite{blalock_what_2020}}                                                              & Down                          & \textcolor[RGB]{65,174,61}{\checkmark}                                  & \textcolor[RGB]{65,174,60}{\checkmark}\tnote{\tiny{4}}                                         & \textcolor[RGB]{65,174,60}{\checkmark}                                    & \textcolor[RGB]{237,90,101}{×}                                        & \textcolor[RGB]{65,174,61}{\checkmark}                                     \\ \cline{2-9}
                                                                                                       & \multicolumn{2}{c|}{Quantization\cite{lin_fixed_2016}}                                                       & Down                                                                                                                                    & \textcolor[RGB]{65,174,60}{\checkmark}                       & \textcolor[RGB]{65,174,60}{\checkmark}\tnote{\tiny{4}}                  & \textcolor[RGB]{65,174,60}{\checkmark}                                   & \textcolor[RGB]{237,90,101}{×}                                                     & \textcolor[RGB]{237,90,101}{×}                                                                                                                               \\ \hline
            \end{tabular}
            \begin{tablenotes}
                \item[1] \textit{Loosely coupled} refers the method is not majorly relying on specific model architecture or task. \textit{Post-training configurable} means the compression rate can be modified without further training. \textit{Real-time configurable} means the compression rate can be modified during the inference runtime.  \textit{Fine-grained Tuning} means the method can achieve arbitrary compression instead of selecting from limited choices.
                \item[2] They meet the prerequisites for implementing these features. Support is possible if our proposed method is applied.
                \item[3] Sparse MoE is more often known as an efficient up scaling method, but created efficient models can be considered as down scaled models.
                \item[4] Backbone training after pruning or quantization is also common, but not necessary.
            \end{tablenotes}

        \end{threeparttable}}\\
\end{table*}
{\cb Model compression and scaling refer to reducing model costs with acceptable performance loss or improving model performance efficiently with increased costs. Most compression methods produce static, immutable models with a fixed compression rate once deployed. Dynamic models, however, can adapt to individual input samples \cite{han_dynamic_2022}. The dynamic nature of these models offers a new dimension for compression and scaling, providing difficulty-aware compression and configurable compression ratios through a dynamic computation graph. We reviewed and categorized model compression and scaling techniques, including both static methods and those utilizing dynamic models, summarizing the primary distinctions in \cref{tab:related-work}.}

\subsection{Compression \& Scaling for General Models}
\noindent \textbf{Architecture Scaling} is an integral part of modern model architectures, allows for the scaling of architectures into different versions \cite{he_deep_2016}\cite{tan_efficientnet_2019} by modifying the number of repeating components. However, this scaling is coarse-grained, and each version of the model must be trained from scratch. \\
\textbf{Quantization\cite{lin_fixed_2016}}  is a prevalent compression technique that uses fewer bits to represent network weights or activations. Quantization cannot reduce the number of operations but can achieve some speedups with specific hardware supports because quantized variables theoretically only require simpler computing units\cite{duModelQuantizationHardware2024}. \\
\textbf{Pruning\cite{blalock_what_2020}\cite{lin_filter_2021}}  is another common approach that removes redundant weights from a trained neural network. However, pruning usually introduces sparsity which may be inefficient for hardware execution. \\
\textbf{Knowledge distillation\cite{hinton_distilling_2015}} is a method to create a smaller model with the teaching of a large model. Unlike Quantization or Pruning which can be applied to both pre-training and post-training phases, Knowledge distillation necessitates a massive training to teach the small model with large model's predictions.

\subsection{Compression \& Scaling Methods Utilizing Dynamic Models}
Dynamic models make it possible to allocate computational resources based on the complexity of each input sample. The \textbf{Early Exit} strategy that enables ``easy'' samples to exit at shallow layers, avoiding the execution of deeper network segments. This concept was initially introduced by BranchyNet\cite{teerapittayanon_branchynet_2016} with confidence-based exits, where computation terminates if the confidence of a classifier exceeds a threshold. Subsequent studies such as FastBERT \cite{liu_fastbert_2020} and PersEPhonEE\cite{leontiadis_its_2021} have extended this confidence-based method across various applications. In these methods, the exit propensity is controlled by external thresholds, thereby enabling post-training tuning. Alternatively, some research uses policy networks rather than thresholds to control exits \cite{chen_learning_2020}\cite{dai_epnet_2020}. These policy networks are trained for specific compression targets, offering improved performance, albeit with less flexibility after training completion. Other work, such as MSDNet\cite{hong_panda_2021},  MSNet\cite{hang_msnet_2023}, DVT\cite{wangNotAllImages2021} and CF-ViT\cite{chenCFViTGeneralCoarsetoFine2022}, achieves high performance by enabling feature reusing. Compared to efficient static models, these approaches offer efficiency by differentiating sample difficulties but necessitate task-specific designs. i.e. these approaches cannot be used to retrofit scalability into an existing base model network. Apart from early exits, which skip all subsequent layers, \textbf{Layer Skipping}\cite{wang_skipnet_2018} is an idea that skips some intermediate layers while retaining the remaining model segments. There are also strategies involving \textbf{Channel Skipping}\cite{lin_runtime_2017}, which omits only certain parts of a layer during runtime. Designing and training models that can adapt to skipped intermediate parts present significant challenges, making these methods less prevalent than the early exit approach. \textbf{Sparse Mixture of Experts (MoE)\cite{fedusSwitchTransformersScaling2022,riquelmeScalingVisionSparse2021,daxbergerMobileVMoEsScaling2023}} is another wide-sued model scaling technique. It employs multiple experts (sub-models in parallel) but adaptively activate only some of them during inference. Typically, experts in MoE implementations possess equal complexity,making Sparse MoE cannot address sample difficulties by varying the amount of computation. Nevertheless, this approach significantly expands model capacity by substantially increasing the parameter count while maintaining consistent computational costs. 

\section{Inference with Early Exits}
\label{sec:system-description}
We start illustrating our system with the inference of a model with multiple early exits. Before the inference, we attach multiple extra networks to a base model as exit points and generate various configurations to coordinate all exits. We describe how to achieve these in \cref{sec:design} and \cref{sec:config-search}. \\
{\cb 
DyCE reduces overall computation by dynamically routing samples based on their difficulty. When inference reaches an exit point, the process terminates if the model's confidence exceeds a predefined threshold. Otherwise, computation continues to the next backbone segment. This approach allocates more resources to difficult inputs, through trigger later exits, while easy inputs are processed through early exits. Additionally, DyCE can switch configurations at runtime to adjust the exit ratio at a macro level. If more accuracy is requested, more samples will be routed to later exits with more computation. Conversely, for low computational cost is more important, samples will exit earlier during inference.}
\subsection{Runtime Architecture}
To illustrate the run-time architecture of DyCE, we divide a pre-trained network (i.e. the base model), e.g., ResNet into different segments, and exit networks are attached to each of them. These exits are trained and grouped into various configurations to obtain the required trade-off. These ideas are now formally defined in the following sub-sections.

\begin{figure}[h]
    \centering
    \includegraphics[width=\linewidth]{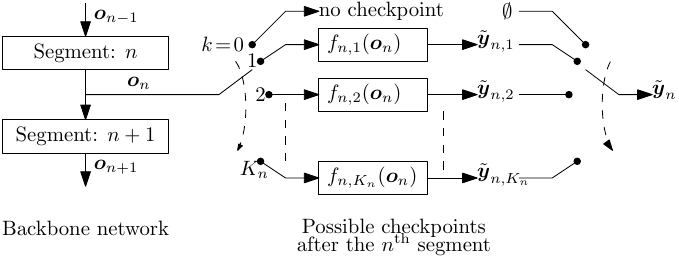}
    \caption[]{Backbone segments and attached exits. At run-time one of $K_n$ possible exits will be applied to the output of the $n^\text{th}$ segment.}
    \label{fig:K_checkpoint_after_nth_segment}
\end{figure}
\subsubsection{Backbone Segments}
A deep neural network can usually be considered as two parts: a backbone for feature extraction and a small network at the end for computing the final output. The backbone part is typically stacked as multiple dividable layers. Hence, we consider that the backbone of a pre-trained network is divided into $N$ segments. The nature of this segmentation could be fine or granular, and we denote the output of the $n^\text{th}$ segment (and the input to $(n+1)^\text{th}$) as $\boldsymbol{o}_n$ as shown in \cref{fig:K_checkpoint_after_nth_segment}.
\subsubsection{Exit}
At the $n^\text{th}$ segment output we can attach one of $K_n$ relatively simple  exit functions, indexed by $k$, whose task is to make an early estimate of the output vector:
\begin{align}
    \tilde{\boldsymbol{y}}_{n,k} & = f_{n,k}\!\left(\boldsymbol{o}_n\right)\label{eq:def_tilde_y_n_k}
\end{align}
where $\tilde{\boldsymbol{y}}_{n,k}$ has a format compatible\footnote{Note that we define that  $\tilde{\boldsymbol{y}}_{n,k}$ is compatible with the original network's output but not necessarily in the exact same format. It could, for example, be shorter, providing classification results for a subset of classes or perhaps some clustering of classes.} with the original network's output. As illustrated in \cref{fig:K_checkpoint_after_nth_segment}, we provide multiple exit candidates at the same position. The configuration searching algorithm (discussed in Sec.5) will identify the best one for a given compression preference, and only that one will be loaded during the inference. In our experiments, we use basic multilayer perceptron (MLP)\cite{gardner_artificial_1998} with different layers and neurons, but they can be any designs align with the definition above. For ease of description, we consider the pre-trained small network at the end of the original backbone network as being another exit, denoted  $f_{N,1}$. Any input sample that does not exit early will eventually pass through this final exit. Each of the exit function is designed and trained offline independently of any other and are fixed at run-time.

\subsubsection{Confidence and Threshold}
The exit confidence is an estimate of the correctness of the result. For classification problems, we take this confidence to be the maximum value of the predicted class probability, i.e., $max(\tilde{\boldsymbol{y}}_{n,k})$.
In our system, we apply a threshold, $t$, to this confidence value in order to make an early-exit decision. A high threshold limits an exit such that only highly confident samples exit early, and a lower threshold will allow more samples to exit early. Accordingly, the threshold, $t$, can be adjusted to obtain different complexity/accuracy trade-offs as will be discussed in \cref{sec:config-search}.
\subsubsection{Exit Group and Run-time Algorithm}
Our system uses pre-defined exit configuration groups, each denoted as $G$, that enumerates which exit functions and what threshold values are to be used at the output of each segment. Specifically we define an exit configuration, $G=\{\boldsymbol{k},\boldsymbol{t}\}$, comprising a length $N$ ordered list of function indices, $\boldsymbol{k}$, and thresholds, $\boldsymbol{t}$, such that the exit to be used after the $n^\text{th}$ segment is $f_{n,k_n}$, i.e. $\boldsymbol{\tilde{y}}_{n} \triangleq \boldsymbol{\tilde{y}}_{n,k_n}$ and the corresponding confidence is $\max(\tilde{\boldsymbol{y}}_n)$. This confidence is then compared against the threshold $t_{n}$ in order to make an early-exit decision.
If the confidence is not less than the threshold, i.e.  $\max(\tilde{\boldsymbol{y}}_n) \ge t_n$, the computation is terminated (early exited) and the $\tilde{\boldsymbol{y}}_n$ vector is returned. %
Note that we set $k_N=1$ and $t_N=0$ forcing the last exit to always be $f_{N,1}$, and forcing all samples exit eventually.
This run-time algorithm is summarized in the \cref{alg:runtime_algorithm}.
\begin{algorithm}
    \KwIn{Exit configuration, $G=\{\boldsymbol{k},\boldsymbol{t}\}$, where $\boldsymbol{k}$ and $\boldsymbol{t}$ and length $N$ vectors.}
    \KwOut{The vector of class probabilities: $\hat{\boldsymbol{y}}$}
    \For{$n=1 \rightarrow N$}
    {
        $k = k_n$ and $t = t_n$\\
        Execute $n^\text{th}$ segment of backbone network $\rightarrow \boldsymbol{o}_n$\\
        \If{$k\ne 0$}{
            $\tilde{\boldsymbol{y}} = f_{n,k}(\boldsymbol{o}_n)$\\
            \If{$\max(\tilde{\boldsymbol{y}}) \ge t$} {
                $\hat{\boldsymbol{y}} \leftarrow \tilde{\boldsymbol{y}}$\\
                Exit early.\\
            }
        }
    }
    \caption{Run-time Inference Algorithm}\label{alg:runtime_algorithm}
\end{algorithm}

\section{Design and Training}
\label{sec:design}
\subsection{Exit Design}

\subsubsection{Network}
The design of an exit must consider the trade-off between complexity and performance. Although more complex exit-ing functions have higher stand-alone performance, they would add more computational overhead and can affect the system's overall performance. One of the possible forms of an exit is the MLP. However, any functions that can generate a compatible output are eligible exiting functions. Meanwhile, since every exit is independent of others, there is no requirement for all exits to have the same architecture. They can be designed and trained separately. It is also possible to design multiple candidate exits for the same position. The searching algorithm (discussed in \cref{sec:config-search}) can find the most suitable one for different targets. In this paper, we use small MLPs as exits to generate results in the same (not just compatible as mentioned previously) format as $f_{N,1}$. We also employ an average pooling layer before the first MLP layer to reduce the feature map's height and width to 1.
\subsubsection{Feature Aggregation}
The input to early exits is from the hidden layer outputs of the original model. However, these intermediate features usually have large dimensions. Directly feeding these features into an early-exit network will result in significant computations. Therefore, for employing a neural network as the exit function, a feature aggregation layer is usually required as the first layer to extract and compress the information from the raw feature map. Convolution layers, pooling layers, MLP layers or any other network that can reduce the feature map size can be used as a feature aggregator. However, feature aggregation layers should be as simple as possible to save resources for the exit network. Our implementation uses average pooling as the feature aggregator to reduce the width and height of feature maps to 1.

\subsection{Training}
\label{sec:ckpt-train}
During the training process, the original backbone network is frozen, and only the exit network is trained.
Freezing the original model ensures the system can restore full performance at any time since the original model is left untouched. It also ensures all exits are independent of each other and the original network because they do not share any trainable network regions. Since the training of each exit is isolated, multiple exits can be trained together or separately with the same or different training recipes. Using the same training recipe as the original network is the simplest way. However, in our experiment, we use the soft cross-entropy loss, making every exit network mimic the output of the original network. For the exit at $n^{th}$ position, its loss function is denoted as:
\begin{align}
    L_n & =  CE (\tilde{\boldsymbol{y}}_n,\tilde{\boldsymbol{y}}_N)                                                                                                                   \\
        & =-\sum_{i=1}^{N_{class}} log\frac{e^{\tilde{\boldsymbol{y}}_n^{(i)}}}{\sum_{i^\prime=1}^{N_{class}}e^{\tilde{\boldsymbol{y}}_n^{(i^\prime)}}}\tilde{\boldsymbol{y}}_N^{(i)}
\end{align}
This approach is inspired from knowledge distillation\cite{hinton_distilling_2015}, as
a part of the backbone network, along with an exit network, can be considered a smaller version of the original model. Compared with the regular cross-entropy loss, this approach can alleviate the over-fitting issue by preventing resource-limited exits pursuing hard labels. This will also allow building or fine-tuning the DyCE system without a labelled dataset if the backbone is pre-trained. Application-specific data can be used to boost this system without labelling them.

\subsection{Ensemble of Exits}
\label{sec:coop-of-ckpt}
\begin{figure}[htbp]
    \centering
    \includegraphics[width=1\linewidth]{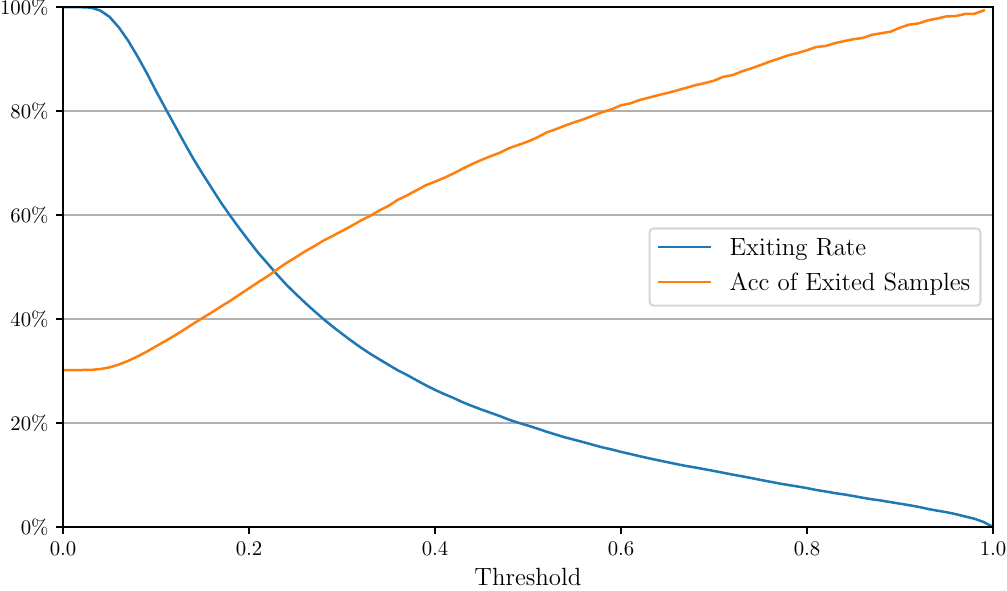}
    \caption{Accuracy of predictions at the $1^{st}$ exit (located at $1^{st}$ skip connection's output) of ResNet$_{\mathit{50}}$ for ImageNet. As the threshold increases, fewer samples exit at this position while the accuracy is rising.}

    \label{fig:performance-with-thres}
\end{figure}

The standalone performance of early attached exits might not be high because the preceding backbone segments (layers) are not sufficient in number and were not originally designed and trained to support this exit. However, by using a group of exits in a cooperative fashion, we significantly reduce incorrect predictions.
\cref{fig:performance-with-thres} shows the accuracy of predictions at an early exit implemented on ResNet$_{\mathit{34}}$. As illustrated, fewer samples exit at higher thresholds but with better accuracies.
This observation indicates that, though stand-alone accuracy at early exits may not be high enough, their high-confidence predictions (at higher thresholds) are very reliable. In a group of early-exits with high thresholds, samples must be exited by either a highly confident exit or by the final exit.
This way, high overall accuracy can be maintained while reducing computational complexity for those samples which exit early.
For varying the model performance in real-time, \cref{sec:config-search} discusses an algorithm to find the most suitable confidence range for every exit in such a way that the overall complexity is reduced for a specific performance target.

\section{Configuration Searching}
\label{sec:config-search}
The configuration of deployed exits should be optimized to achieve specific performance or complexity targets.

However, the search space is vast as the number of possible exit combinations are $\prod_{n=1}^{N-1}(K_n +1)$, and each of them has thresholds. It is extremely difficult to go through every possibility to find the optimal configuration. Therefore, one contribution of this work is to provide search algorithms for identifying which exit configuration should be used under a specific constraint. Our technique also allows changing the constraints in real time to achieve adjustable trade-offs.

\subsection{Performance metrics}
Configuration searching should consider two sets of goals: Model performance and Complexity. In this work, we equate them to the overall model accuracy and the number of MAC operations relative to the original model.
\subsubsection{Relative Overall Accuracy}
The overall accuracy is computed by aggregating predictions from all exits. %
Assume that we are working on a model with $N$ segments and a dataset of $M$ samples. By executing the inference algorithm on the dataset, for every sample, all exit outputs in a configuration $G= \{\boldsymbol{k},\boldsymbol{t} \}$ is recorded as $\tilde{\boldsymbol{y}}_{n,k_n}$ and the predicted probability of $i^{th}$ class for $m^{th}$ sample is represented as $\tilde{\boldsymbol{y}}^{(m,i)}_{n,k_n}$.
We then define $W_n(\boldsymbol{k},\boldsymbol{t})$ as indices of samples that have confidence higher than the corresponding threshold $t_n$ at the $n^{th}$ exit. $U_n(\boldsymbol{k},\boldsymbol{t})$ is $W_n(\boldsymbol{k},\boldsymbol{t})$ excluding any previously exited samples, i.e., indices of samples which are exited at $n^{th}$ position. $V_n(\boldsymbol{k},\boldsymbol{t})$ is a subset of $U_n(\boldsymbol{k},\boldsymbol{t})$ where every prediction is correct. These variables are defined mathematically in Eqn. \eqref{eqn:wn}, \eqref{eqn:un} and \eqref{eqn:vn}.
\begin{align}
    W_n(\boldsymbol{k},\boldsymbol{t}) & = \{m: \mathop{max}_{i}(\tilde{\boldsymbol{y}}^{(m,i)}_{n,k_n})\ge t_n\}\label{eqn:wn}                                                          \\
    U_n(\boldsymbol{k},\boldsymbol{t}) & = W_n(\boldsymbol{k},\boldsymbol{t})\setminus {\textstyle \bigcup_{n^\prime=1}^{n-1} W_{n^\prime}}(\boldsymbol{k},\boldsymbol{t})\label{eqn:un} \\
    V_n(\boldsymbol{k})                & = \{m: \mathop{argmax}_{i}(\tilde{\boldsymbol{y}}^{(m,i)}_{n,k_n})=\boldsymbol{y}^{(m,i)}\}\label{eqn:vn}
\end{align}
where the maximization in Eqn. \eqref{eqn:vn} are done over all possible classes (indexed by $i$). The number of samples that reach and exit from the $n^{th}$ exit of configuration $G=\{\boldsymbol{k},\boldsymbol{t}\}$ can be counted as $E_{n}(k,t)$, and the number of these that are correct can be counted as $EC_{n}(k,t)$.
\begin{align}
    E_n(\boldsymbol{k},\boldsymbol{t})  & = \left | U_n(\boldsymbol{k},\boldsymbol{t})\right |                         \\
    EC_n(\boldsymbol{k},\boldsymbol{t}) & = \left | U_n(\boldsymbol{k},\boldsymbol{t})\cap V_n(\boldsymbol{k})\right |
\end{align}
where $\left | S \right |$ means the cardinality of a set $S$.\\
Then the accuracy is measured by the corrected predictions over the total, i.e., \begin{align}A_n(\boldsymbol{k},\boldsymbol{t})=\frac{EC_n(\boldsymbol{k},\boldsymbol{t})}{E_n(\boldsymbol{k},\boldsymbol{t})}\end{align}

Now we define the overall accuracy as:
\begin{align}
    A(\boldsymbol{k},\boldsymbol{t}) = \frac{1}{M\cdot A_{ori}}\sum_{n=1}^{N} \left | E_n(\boldsymbol{k},\boldsymbol{t})  \right | A_n(\boldsymbol{k},\boldsymbol{t})
\end{align} where $A_{ori}$ is the accuracy of the original model. This is a normalized factor making $A(\boldsymbol{k},\boldsymbol{t}) \in [0,1]$.

\subsubsection{Relative Run-time Complexity}
This paper considers the number of MAC operations as a proxy for complexity. Moreover, we normalize the number of MAC operations in a given functional block with respect to the total number of MAC operations in the original network. We define the following normalised complexity measures:
\begin{inparaenum}[1)]
    \item Let $\text{S}_n$ be the normalized complexity of the $n^\text{th}$ segment alone. Accordingly $\sum_{n=1}^N\text{S}_n=1$.
    \item Let $\Delta_{n,k}\ge0$ be the complexity of the exit function $f_{n,k}$. Note that $\Delta_{n,0}=0$ for all $n$ as this is the ``no exit'' i.e. exit disabled, option.
\end{inparaenum}
Then the average run-time complexity of a configuration is defined as:
\begin{align}
    C(\boldsymbol{k},\boldsymbol{t}) & = \frac{1}{M}\sum_{n=1}^{N} E_{n}(k_n,t_n) \sum_{n^\prime=1}^{n}\left(S_{n^\prime} + \Delta_{n^\prime,k_{n^\prime}}\right)
\end{align}

\subsection{Generating configurations}
\subsubsection{Optimization Target}
The model compression needs to take both performance and complexity into consideration. Our proposed dynamic compression method is designed to generate a series of configurations with respect to the relative importance of performance and complexity. We summarise both factors with a parameter $\lambda$ into the following target function:
\begin{align}
    f(\boldsymbol{k},\boldsymbol{t},\lambda) & \triangleq \lambda\left(1-A(\boldsymbol{k},\boldsymbol{t})\right) + (1-\lambda)C(\boldsymbol{k},\boldsymbol{t}) \label{eq:def_average_cost_func}
\end{align}
Where $0 \le \lambda\le 1$ is a user-controlled parameter to adjust the relative importance of accuracy and complexity, i.e., as $\lambda \rightarrow 1$, then the object becomes system classification error rate $=1-A(G)$ and so a minimization process would maximize the accuracy with no consideration to the complexity cost. Conversely, as $\lambda \rightarrow 0$, solutions with minimal complexity would be found at the cost of accuracy. This function will be used as the target of a minimization algorithm.\\

\subsubsection{Searching for configurations}
The configuration search problem is a non-differentiable and non-continuous multi-variable optimization problem, making it exceedingly difficult to devise algorithms that directly optimize $f(\boldsymbol{k},\boldsymbol{t},\lambda)$. Thus, we reformulate this problem as a single-variable problem by only allowing modifications to an existing configuration $G={\boldsymbol{k},\boldsymbol{t}}$ at one position. Consequently, the objective function transforms into:

\begin{align}
    f_{\boldsymbol{k},\boldsymbol{t},\lambda,n}(k_{c},t_{c})~\triangleq ~ & \lambda\left(1-A_{\boldsymbol{k},\boldsymbol{t},n}(k_{c},t_{c})\right) \\\nonumber &+ (1-\lambda)C_{\boldsymbol{k},\boldsymbol{t},n}(k_{c},t_{c})
\end{align}

Here, $(k_{c},t_{c})$ denotes the latent action at the $n^{th}$ exiting position (At $n^{th}$ position, exit $k_{c}$ is called to make a prediction and produce the confidence. The computing is terminated if the confidence is greater than $t_{c}$.). {\cb Empirically, the relationship between $A_{\boldsymbol{k},\boldsymbol{t},n}(k_{c},t_{c})$ and $C_{\boldsymbol{k},\boldsymbol{t},n}(k_{c},t_{c})$ is like \cref{fig:selectable-example}. When more computation is used, the accuracy is improved and it will be more difficult to further improve the accuracy as the computation increases. i.e. We have:
\begin{align}
    \frac{\partial A(k_{c},t_{c})}{\partial C(k_{c},t_{c})} > 0~~&and~~\frac{\partial^2 A(k_{c},t_{c})}{\partial C^2(k_{c},t_{c})} < 0\\
    \lim_{t_c \to 0} \frac{\partial A(k_{c},t_c)}{\partial C(k_{c},t_c)}=\infty~~&and~~\lim_{t_c \to 1} \frac{\partial A(k_{c},t_c)}{\partial C(k_{c},t_c)}=0    
\end{align}
That means, $\frac{\partial A_{\boldsymbol{k},\boldsymbol{t},n}(k_{c},t_{c})}{\partial C_{\boldsymbol{k},\boldsymbol{t},n}(k_{c},t_{c})}$ is positive and monotonically decreasing. Therefore, the equation $f^{\prime}_{\boldsymbol{k},\boldsymbol{t},\lambda,n}(k_{c},t_{c})=0$ has and only has one solution. i.e. $f_{\boldsymbol{k},\boldsymbol{t},\lambda,n}(k_{c},t_{c})$ has and only has one minimum. Hence, the object fuction is convex and the searching algorithm can converge to the global minimal. }\\

Once $f_{\boldsymbol{k},\boldsymbol{t},\lambda,n}(k_{c},t_{c})$ can be minimized, we traverse different layers and exit types to ascertain the optimal action. This action is then applied to the existing configuration. This process is repeated until the round limit is reached, or no additional action can enhance the current configuration. This search method is encapsulated in our proposed \textit{Iterative Search}, as depicted in \cref{alg:circular_search_algorithm}. Although this algorithm employs a greedy approach, which does not guarantee convergence to the optimum, it effectively narrows the search space to an acceptable range.

\begin{algorithm}[h]
    \KwIn{$\lambda$}
    \KwOut{Exit configuration, $G=\{\boldsymbol{k},\boldsymbol{t}\}$.}
    Initialize $\boldsymbol{k}$ and $\boldsymbol{t}$ to be $N\times 1$ vectors of zeros, i.e. no exits by default.\\
    Initialize cost metric, $f_{min}= f(\boldsymbol{k},\boldsymbol{t},\lambda)$.\\
    \While{$\mathrm{True}$} {
    $\mathrm{Candidate} \leftarrow \mathrm{None}$\\
    \For{$n=1 \rightarrow N$}	{
    \For{$k_c=0 \rightarrow K_{n}$}	{
    $\{t_c,~f_c\} \leftarrow \mathop{min}_{t_c} f_{\boldsymbol{k},\boldsymbol{t},\lambda,n}(k_{c},t_{c})$ \\
    \If{$f_c < f_{min}$}{
        $\mathrm{Candidate} \leftarrow \{n, k_c, t_c\}$\\
        $f_{min} \leftarrow f_c$
    }
    }
    }
    \uIf{$\mathrm{Candidate} ~\text{is not} ~\mathrm{None}$}{
        $n,~k_c,~t_c \leftarrow \mathrm{Candidate}$\\
        $t_n \leftarrow t_c$ \hspace{0.5cm} $k_n \leftarrow k_c$ \\
    }
    \Else{
        \textbf{break}
    }

    }
    \caption{Iterative Search algorithm}\label{alg:circular_search_algorithm}
\end{algorithm}
In \cref{alg:circular_search_algorithm} $\mathop{min}_{t_c} f_{\boldsymbol{k},\boldsymbol{t},\lambda,n}(k_{c},t_{c})$ is a bounded minimization algorithm, e.g., golden section search to find a $t_c \in (0,1)$ such that the function value is minimized to $f_c$.
\\
However, the iterative search algorithm will still be time-consuming when dealing with large datasets. As an alternative, a simpler and more straightforward substitution is proposed, i.e. the single-pass search algorithm. This method involves only one pass from the first to the last segment. In this algorithm, the search process starts with an assumed empty configuration. Then we begin with the first possible exiting position ($n=1$) and add the option to help the configuration achieve the lowest $f_{c}$. With this action taken into use, we then consider the next exit position until $n=N$, i.e., we make a single-pass through all possible positions hence the name. The single-pass algorithm dramatically saves the searching time, and our experiments indicate it do not have a significant impact on results.
\begin{algorithm}[htbp]
    \KwIn{$\lambda$}
    \KwOut{Exit configuration, $G=\{\boldsymbol{k},\boldsymbol{t}\}$.}
    Initialize $\boldsymbol{k}$ and $\boldsymbol{t}$ to be $N\times 1$ vectors of zeros, i.e. no exits.\\
    Initialize cost metric, $f_{min}= f(\boldsymbol{k},\boldsymbol{t},\lambda)$.\\
    \For{$n=1 \rightarrow N$}	{
    \For{$k_c=0 \rightarrow K$}	{
    $\{t_c,~f_c\} \leftarrow \mathop{min}_{t_c} f_{\boldsymbol{k},\boldsymbol{t},\lambda,n}(k_{c},t_{c})$\\
    \If{$f_c < f_{min}$}{
        $f_{min} \leftarrow f_c$ \hspace{0.5cm} $t_n \leftarrow t_c$ \hspace{0.5cm} $k_n \leftarrow k_c$\\
    }
    }
    }
    \caption{Single-pass search algorithm}\label{alg:single_pass_top_level_search_algorithm}
\end{algorithm}

\subsection{Configurations for different targets}
\label{sec:lambda}

The result of the search algorithm is a specific configuration that exhibits a trade-off between complexity and performance, determined by the parameter $\lambda$. This search process can be replicated with diverse values of $\lambda$ to derive additional configurations with varying trade-off characteristics. A device executing the model for inference can store multiple pre-defined configurations and toggle between them at runtime. This capability facilitates dynamic model compression depending on the real-time requirements. The interval between multiple $\lambda$ values can be kept minimal to enable fine-grained performance tuning.
\section{Simulation}
\label{sec:simulation}
\subsection{Base Model and Dataset}
In 2016, He et al. presented ResNet\cite{he_deep_2016}, which is a residual learning framework to train very deep networks for computer vision tasks. After that, many studies proved that ResNet and its residual structure have a successful performance on various computer vision tasks. The core element of ResNet is the residual block which consists of two or three cascaded convolutional layers with a shortcut. Many of today's state-of-the-art deep CNNs still follow this structure. {\cb 
This paper validates the proposed early-exit algorithm on ResNet \cite{he_deep_2016}, ConvNeXtv2 \cite{woo_convnext_2023}, and DaViT \cite{ding_davit_2022}. The observation is applicable to other deep neural networks as well. ImageNet-1k \cite{russakovsky_imagenet_2015}, a classic dataset for computer vision and artificial intelligence, contains 1.28 million training images and 50,000 validation images of size 224x224. It has been widely used as a benchmark for image classification, and results on the ImageNet-1k validation set are considered a reference for model generalizability in practice. This paper simulates variants of the three backbone architectures at different depths on ImageNet-1k to demonstrate the effectiveness of the proposed method.}

\subsection{Model Implementation}
The backbone part of our model implementation for the ImageNet dataset follows the repository by PyTorch\cite{team_ResNet_nodate}. We attach exits to the output of each residual block in {\cb the base model. For example, }ResNet$_{\mathit{34}}$ and ResNet$_{\mathit{50}}$ have 16 positions for exits, including the last exit (the head of the original models). There are 50 exit points for ResNet$_{\mathit{152}}$ since it has more layers. The main purpose of this work is not to compete on performance; we employ five different types of plain MLP, from small to large, for every position except the last one. These five types are MLPs of 1-layer, 3-layer with 500 neurons in each hidden layer, 3-layer with 1000 neurons, 5-layer with 500 neurons, and 5-layer with 1000 neurons. The search algorithm will select either none or one of the options for each position in a configuration.

\subsubsection{Training details}
\label{sec:sim:imagenet:train}
The training of the exits is based on an existing pre-trained model. The backbone network is freezed and only the exits will be trained. The forward propagation will go through the whole model, while the backward propagation works only on each exit. The data augmentation follows the same strategy as ImageNet pre-training, and the optimizer is Adam with a learning rate $lr=3\times10^{-4}$. The loss function is soft cross entropy as mentioned in \cref{sec:ckpt-train}.

\subsection{Performance Evaluation}

\begin{figure*}[ht]
    \centering
    \includegraphics[width=1\linewidth]{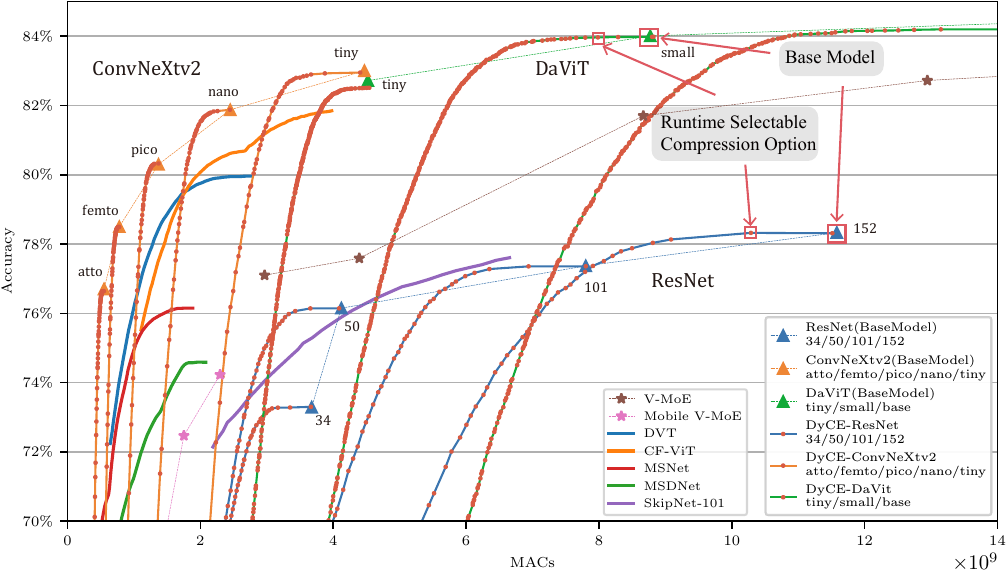}
    \caption{{\cb MACs and Accuracy of the proposed method, base models and related work. Original ResNet, ConvNeXtv2 and DaViT variants are represented as triangles, the compressed version is denoted by the curve attached to each of them. The performance curve of related work is also plotted for reference. These models can not be configured in runtime or adaptive to new base models. However, DyCE can switch to different configurations (denoted by red dots) in runtime and adapt models with different architectures.}}
    \label{fig:performance-curve}
\end{figure*}

In this study, we present an evaluation of performance to substantiate the effectiveness of our proposed method.  \cref{fig:performance-curve} provides a comparative visualization of the performance and computational requirements between our proposed approach and other methods. Our approach is validated with base models such as ResNet\cite{he_deep_2016} and ConvNeXtv2\cite{woo_convnext_2023}. Each point on the depicted curve signifies an available configuration, providing the option for the user to select in real-time.

\begin{table}[h]
    \centering
    \caption{{\cb Comparison among DyCE enabled models and other models with configurable efficiency. The accuracy on the ImageNet validation set is reported. The DyCE enabled models are marked in bold.}} \label{tab:quantative-results}

    \resizebox{0.8\linewidth}{!}{
        \begin{threeparttable}[b]
            \begin{tabular}{l|ll}
                \hline
                Model                                                        & GMACs          & Acc              \\ \hline
                DaViT$_\mathit{base}$\cite{ding_davit_2022}                  & 15.98          & 84.49\%          \\
                ResNet$_{152}$\cite{he_deep_2016}                            & 11.51          & 78.31\%          \\
              \textbf{DyCE DaViT$_\mathit{base}$\cite{ding_davit_2022}}         & \textbf{10.87} & \textbf{83.51\%} \\ \hline
                DaViT$_\mathit{small}$\cite{ding_davit_2022}                 & 8.77           & 84.01\%          \\
                ResNet$_{101}$\cite{he_deep_2016}                            & 7.80           & 77.36\%          \\
                SkipNet\cite{wang_skipnet_2018}                              & 6.65           & 77.61\%          \\
              \textbf{DyCE DaViT$_\mathit{small}$\cite{ding_davit_2022}}        & \textbf{6.50}  & \textbf{83.03\%} \\
                SkipNet\cite{wang_skipnet_2018}                              & 5.64           & 76.61\%          \\ \hline
                DaViT$_\mathit{tiny}$\cite{ding_davit_2022}                  & 4.52           & 82.71\%          \\
                ConvNeXtV2$_\mathit{tiny}$\cite{woo_convnext_2023}           & 4.46           & 82.95\%          \\
                ResNet$_{50}$\cite{he_deep_2016}                             & 4.09           & 76.14\%          \\
                CF-ViT\cite{chenCFViTGeneralCoarsetoFine2022}                & 3.98           & 81.85\%          \\
                SkipNet\cite{wang_skipnet_2018}                              & 3.97           & 73.69\%          \\
                ResNet$_{34}$\cite{he_deep_2016}                             & 3.66           & 73.30\%          \\
              \textbf{DyCE ConvNeXtV2$_\mathit{tiny}$\cite{woo_convnext_2023}}  & \textbf{3.27}  & \textbf{82.03\%} \\
                CF-ViT\cite{chenCFViTGeneralCoarsetoFine2022}                & 3.13           & 80.87\%          \\ \hline
                DVT\cite{wangNotAllImages2021}                               & 2.76           & 79.96\%          \\
                ConvNeXtV2$_\mathit{nano}$\cite{woo_convnext_2023}           & 2.45           & 81.87\%          \\
                SkipNet\cite{wang_skipnet_2018}                              & 2.34           & 67.61\%          \\
                MSDNet\cite{hong_panda_2021}                                 & 2.08           & 74.59\%          \\
              \textbf{DyCE ConvNeXtV2$_\mathit{nano}$\cite{woo_convnext_2023}}  & \textbf{1.97}  & \textbf{80.91\%} \\
                CF-ViT\cite{chenCFViTGeneralCoarsetoFine2022}                & 1.89           & 77.86\%          \\
                MSNet\cite{hang_msnet_2023}                                  & 1.89           & 76.15\%          \\
                DVT\cite{wangNotAllImages2021}                               & 1.73           & 78.98\%          \\
                MSDNet\cite{hong_panda_2021}                                 & 1.63           & 73.61\%          \\
                ConvNeXtV2$_\mathit{pico}$\cite{woo_convnext_2023}           & 1.37           & 80.32\%          \\
                DVT\cite{wangNotAllImages2021}                               & 1.31           & 76.15\%          \\
                CF-ViT\cite{chenCFViTGeneralCoarsetoFine2022}                & 1.27           & 71.85\%          \\
                MSDNet\cite{hong_panda_2021}                                 & 1.23           & 70.62\%          \\
                MSNet\cite{hang_msnet_2023}                                  & 1.20           & 75.16\%          \\
              \textbf{DyCE ConvNeXtV2$_\mathit{pico}$\cite{woo_convnext_2023}}  & \textbf{1.19}  & \textbf{79.38\%} \\ \hline
                MSNet\cite{hang_msnet_2023}                                  & 0.89           & 72.16\%          \\
                DVT\cite{wangNotAllImages2021}                               & 0.88           & 70.06\%          \\
                ConvNeXtV2$_\mathit{femto}$\cite{woo_convnext_2023}          & 0.78           & 78.49\%          \\
                MSDNet\cite{hong_panda_2021}                                 & 0.78           & 64.76\%          \\
              \textbf{DyCE ConvNeXtV2$_\mathit{femto}$\cite{woo_convnext_2023}} & \textbf{0.70}  & \textbf{77.53\%} \\
                MSNet\cite{hang_msnet_2023}                                  & 0.64           & 66.18\%          \\
                ConvNeXtV2$_\mathit{atto}$\cite{woo_convnext_2023}           & 0.55           & 76.64\%          \\
              \textbf{DyCE ConvNeXtV2$_\mathit{atto}$\cite{woo_convnext_2023}}  & \textbf{0.49}  & \textbf{75.65\%} \\
              \hline
            \end{tabular}%
        \end{threeparttable}
    }
\end{table}

{\cb DyCE, as a compression method, effectively conserves computational resources while maintaining minimal performance degradation. \cref{tab:quantative-results} compares DyCE-enabled models to other approaches. Compared to the original versions, DyCE reduces Multiply-Accumulate operations (MACs) by approximately 10\% to 30\% with only a small drop in accuracy. Furthermore, DyCE-enabled models outperform related methods by achieving higher accuracy within a comparable MAC range.}

Further information is visualized in \cref{fig:performance-curve}, where we have plotted the DyCE compression curve for each base model. Each dot on the curve denotes a run-time selectable configuration that enables a different trade-off point between complexity and performance. We employed a step size of 0.01 for $\lambda \in [0,1]$ to generate these curves. The density can be augmented to facilitate almost continuous performance tuning.

DyCE exhibits advantages compared with the native architecture scaling, particularly in regions close to the base model. The DyCE curve for ResNet$_{\mathit{152}}$ initially surpasses the linear trajectory between ResNet$_{\mathit{152}}$ and ResNet$_{\mathit{101}}$ but subsequently falls slightly below the point representing ResNet$_{\mathit{101}}$. Nevertheless, ResNet$_{\mathit{50}}$ augmented with DyCE, consistently outperforms ResNet$_{\mathit{34}}$. These findings underscore the potential of DyCE as a beneficial supplement to conventional static model scaling.

Comparative results from other dynamic scaling methods, such as V-MoE, Mobile V-MoE, DVT, CF-ViT, MSDNet, MSNet, and SkipNet-101, are also depicted in \cref{fig:performance-curve}. These models are task-specific and not runtime configurable, their performance degrade more slowly compared to our plain MLP exits. Given the independence of each exit within DyCE, this challenge can potentially be mitigated by introducing delicately designed exiting networks. Despite the simplicity of the MLPs, DyCE can surpass the efficiency of other methods when suitable base models are chosen. As the proposed framework is decoupled from the base model, it facilitates seamless integration with state-of-the-art models, such as ConvNeXtv2 and {\cb DaViT}. The evaluation results of DyCE in conjunction with ConvNeXtv2 and {\cb DaViT} denote a substantial improvement over competing methods. While other techniques generally require model-specific knowledge for designing exits, they may not be easily applicable to the latest models. Considering the rapid evolution of deep learning, the flexibility offered by DyCE is a valuable feature.

\subsection{Comparison between Searching Algorithms}

We have proposed two search algorithms for generating configurations: the iterative method, which is expected to yield better convergence, and the single-pass method, which offers significantly faster completion. The time complexity of the single-pass algorithm is directly proportional to the number of candidate exits, while the iterative method requires repetitive inspection of every exit, leading to substantially higher complexity. In our experiments, both algorithms have been implemented with GPU acceleration. The single-pass method requires 1.5 minutes to find 100 different configurations for ResNet$_{\mathit{50}}$, whereas the iterative method requires 28.9 minutes for completion. The disparity in time taken becomes more pronounced with an increase in the number of exits. For instance, ResNet$_{\mathit{152}}$, which has 50 exit positions, requires 4.9 minutes and 215.5 minutes to finish using the two respective methods.

\begin{figure}[h]
    \centering
    \includegraphics[width=1\linewidth]{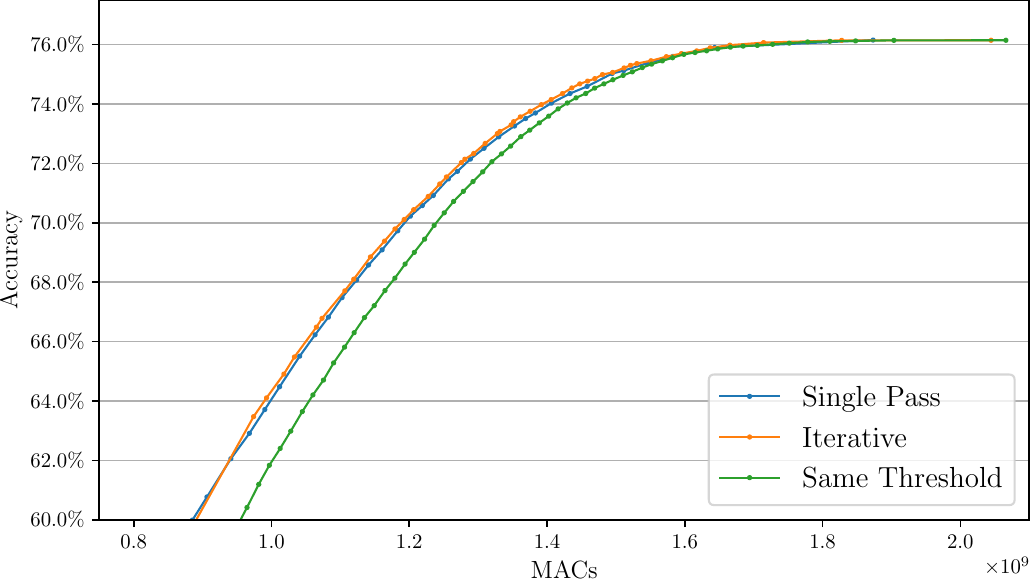}
    \caption{Comparison of three configuration generating methods on ResNet$_{\mathit{50}}$.}
    \label{fig:resnet50-algo}
\end{figure}

Despite the substantial time investment, as depicted in \cref{fig:resnet50-algo}, the iterative method provides a slight edge over the single-pass method. However, considering that configurations only need to be generated once, we recommend utilizing the single-pass method for previewing and debugging purposes and the iterative method for the final generation.

We also compared our proposed algorithms with a baseline method, which selects only one type of exit and applies identical thresholds to all exits. This is the most simplistic approach to make existing early-exit based systems configurable in real time. \cref{fig:resnet50-algo} demonstrates that even when the best exit type is chosen (a 5-layer MLP with 1000 neurons in hidden layers), both of our proposed algorithms exhibit considerable advantages. These results indicate the effectiveness of the search algorithm we propose.

\subsection{Ensemble vs Individual Exits}

\begin{figure}[h]
    \centering
    \includegraphics[width=1\linewidth]{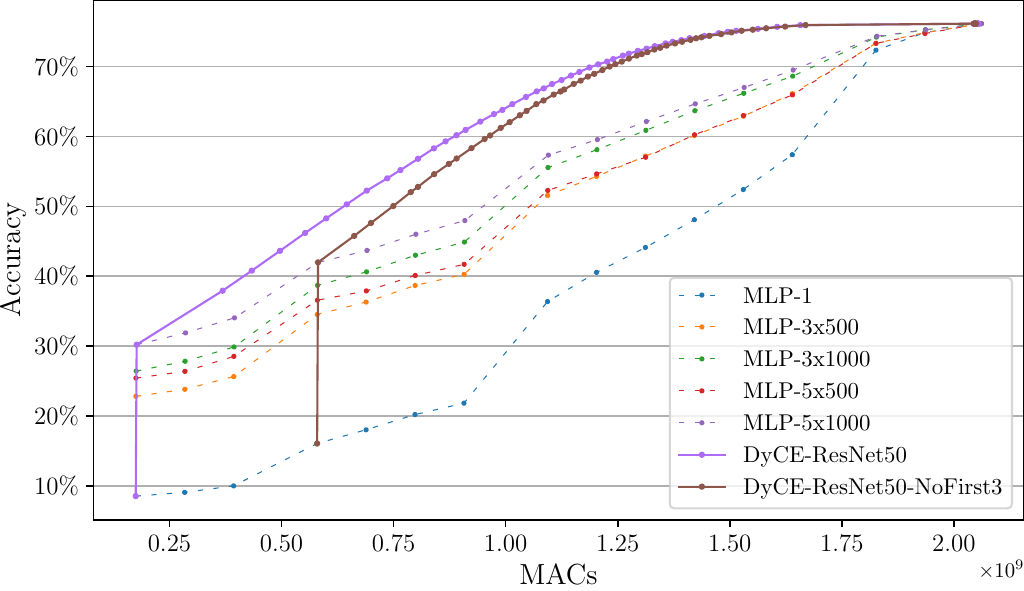}
    \caption{Performance comparison between searched configurations and standalone exits on ResNet$_{\mathit{50}}$. MLP-\underline{A}x\underline{B} refers to An MLP of \underline{A} layers with \underline{B} neurons in all hidden layers.}
    \label{fig:resnet50-sd}
\end{figure}

\cref{fig:resnet50-sd} illustrates the efficacy of utilizing multiple exits collaboratively (represented by the purple line), demonstrating superior performance in comparison to individual exits (represented by the dots in dashed lines). To obtain standalone performance, we emulate a configuration with only one enabled exit and set its threshold to zero, ensuring inference consistently exits at that position. This process is repeated for all potential exits, culminating in the series of dots depicted. As a result, the most optimal type of exit network is ``MLP-5x1000'', but the ensemble of exits significantly outperforms any individual types, including the optimal one. As discussed in \cref{sec:coop-of-ckpt}, the early exit in more trustable when a threshold is applied. In an exiting configuration, most exits except the final one have thresholds, making most exits perform better than their standalone performance. The systematic average computation is the weighted sum of every exit's complexity but the accuracy is higher than the sum of exit's accuracy with the same weight. This observation demonstrates the compression efficiency of early exit groups.

Concurrently, exits at all positions contribute to the overall system performance. The brown line in \cref{fig:resnet50-sd} represents a scenario where the first three positions are disabled. Despite their low accuracy, disabling these exits influences the highest accuracy area, demonstrating that exits at shallow layers have the potential to deal with `easy' samples, and our proposed method can enable them at the right time.

{\cb
\section{Limitations and Future Work}

\label{sec:extension}
This paper demonstrates DyCE with the example of image classification. However, DyCE can be applied to more applications. In this section, we discuss its limitation and the potential of using DyCE in more cases.
}
{\cb \subsection{Constraints and Limitations}

DyCE is a highly flexible framework with notable advantages, but it also has some limitations. During deployment, DyCE requires a dynamic computational graph for inference, which is not well-supported by certain software libraries or hardware accelerators. This requirement may complicate deployment in some scenarios. Additionally, DyCE introduces extra parameters for exit candidates, and the size and number of exits might be restricted in practical implementations. These constraints are common to most dynamic models, and the community is actively seeking solutions. From a development perspective, DyCE assumes that intermediate features from pre-trained backbones are informative enough for prediction. This assumption simplifies system development since the backbone network can then remain unmodified. However, this approach may not achieve optimal performance, as the backbones were not originally designed for this purpose. Co-designing the exits alongside the backbone network could potentially yield better results.

}
\subsection{Besides Image Classification}

{\cb
This paper demonstrates DyCE using an image classification task as an example. However, DyCE can theoretically be adapted to other tasks and models without major barriers. DyCE can be applied to any model that can be divided along its depth, provided that exits are designed to generate candidate predictions in a format compatible with the original output and produce confidence scores for those predictions.

To adapt DyCE to other data types, such as video, audio, or text, the backbone encoder should be replaced accordingly. Encoder architectures for these data types, like RNNs \cite{das_recurrent_2023} and Transformers \cite{vaswani_attention_2017}, are also composed of stacked, dividable blocks, making them suitable for DyCE. When adapting DyCE for tasks beyond classification, the exit component should be designed and trained specifically for the new task. For instance, the exit component for object detection should function as a detector, while for image-to-text tasks, it should act as a caption generator.

}
\subsection{Hierarchical Multi-tasking}
Due to the independence of each checkpoint, this system can be modified to assign different tasks to each checkpoint, such as face recognition, object detection, and image-to-text. These checkpoints with different tasks can be organized in a hierarchical order for specific applications. For the example of the face recognition task, we can assign coarse-grained classifiers to the first few checkpoint positions to determine if there are objects in front of the camera. The next few checkpoints can be fine-grained classifiers to confirm that a human face is in sight. After that, the last few checkpoints can run the regular recognition task. Achieving high performance with a small network is difficult, but it is still possible to get rough answers with limited computation, because the checkpoints attached to the shallow layers can focus on simple but common subtasks. Therefore the majority of the model remains in sleep mode in most cases.
\subsection{Distributed Inference}
The early-exit based model inference is done in a hierarchical manner. Since each exit can generate independent outputs, it is possible to partition the model on different devices for inference. DyCE can be used with model partitioning methods\cite{kimDNNPartitioningFramework2024, liuAdaptiveDNNInference2023} to deploy models on multiple devices. In this situation, DyCE can replace MACs with measured latency while generating configurations. The inference can exit with a very short latency if the initial early-exit network is enough to give an acceptable result. For complex events, the edge device can transmit that sample to fog or cloud nodes for more comprehensive inference. This scheme deals with most samples locally to provide real-time feedback but also tackles complex events by using large networks on remote servers. With DyCE, the overall performance of this heterogeneous system will be estimable and controllable.

\section{Conclusion}
This paper introduces DyCE, a real-time configurable model compression and scaling technique for deep learning models. DyCE simplifies the design process of early-exit-based dynamic compression systems by partitioning the considerations during the design of such systems. It optimizes the cooperation of exits to meet arbitrary compression targets with more efficiency. Furthermore, DyCE introduces a second layer of dynamics to support real-time, fine-grained adjustments to the compression target. This enables applications that employ DyCE to adapt to varying practical demands.

Our experiments validate the effectiveness of the proposed method for image classification tasks employing {\cb vision models}. However, the concept of DyCE is not restricted to these domains and could potentially be extended to other models and applications.
\\[0.5cm]
\textbf{Declaration of generative AI and AI-assisted technologies in the writing process}\\
During the preparation of this work the authors used ChatGPT in order to improve the readablity. After using this tool, the authors reviewed and edited the content as needed and take full responsibility for the content of the publication.

\bibliographystyle{elsarticle-num}
\bibliography{references}

\begin{thebibliography}{10}
\expandafter\ifx\csname url\endcsname\relax
  \def\url#1{\texttt{#1}}\fi
\expandafter\ifx\csname urlprefix\endcsname\relax\def\urlprefix{URL }\fi
\expandafter\ifx\csname href\endcsname\relax
  \def\href#1#2{#2} \def\path#1{#1}\fi

\bibitem{dayarathna_data_2016}
M.~Dayarathna, Y.~Wen, R.~Fan, Data {Center} {Energy} {Consumption} {Modeling}: {A} {Survey}, IEEE Communications Surveys \& Tutorials 18~(1) (2016) 732--794, conference Name: IEEE Communications Surveys \& Tutorials.
\newblock \href {https://doi.org/10.1109/COMST.2015.2481183} {\path{doi:10.1109/COMST.2015.2481183}}.

\bibitem{wu_mixed_2018}
B.~Wu, Y.~Wang, P.~Zhang, Y.~Tian, P.~Vajda, K.~Keutzer, \href{http://arxiv.org/abs/1812.00090}{Mixed {Precision} {Quantization} of {ConvNets} via {Differentiable} {Neural} {Architecture} {Search}}, arXiv:1812.00090 [cs] (2018).
\newline\urlprefix\url{http://arxiv.org/abs/1812.00090}

\bibitem{blalock_what_2020}
D.~Blalock, J.~J. Gonzalez~Ortiz, J.~Frankle, J.~Guttag, \href{https://proceedings.mlsys.org/paper/2020/hash/d2ddea18f00665ce8623e36bd4e3c7c5-Abstract.html}{What is the {State} of {Neural} {Network} {Pruning}?}, Proceedings of Machine Learning and Systems 2 (2020) 129--146.
\newline\urlprefix\url{https://proceedings.mlsys.org/paper/2020/hash/d2ddea18f00665ce8623e36bd4e3c7c5-Abstract.html}

\bibitem{lin_filter_2021}
M.~Lin, L.~Cao, S.~Li, Q.~Ye, Y.~Tian, J.~Liu, Q.~Tian, R.~Ji, Filter {Sketch} for {Network} {Pruning}, IEEE Transactions on Neural Networks and Learning Systems (2021) 1--10\href {https://doi.org/10.1109/TNNLS.2021.3084206} {\path{doi:10.1109/TNNLS.2021.3084206}}.

\bibitem{teerapittayanon_branchynet_2016}
S.~Teerapittayanon, B.~McDanel, H.~Kung, {BranchyNet}: {Fast} inference via early exiting from deep neural networks, in: 2016 23rd {International} {Conference} on {Pattern} {Recognition} ({ICPR}), 2016, pp. 2464--2469.
\newblock \href {https://doi.org/10.1109/ICPR.2016.7900006} {\path{doi:10.1109/ICPR.2016.7900006}}.

\bibitem{kaya_shallow-deep_2019}
Y.~Kaya, S.~Hong, T.~Dumitras, \href{https://proceedings.mlr.press/v97/kaya19a.html}{Shallow-{Deep} {Networks}: {Understanding} and {Mitigating} {Network} {Overthinking}}, in: Proceedings of the 36th {International} {Conference} on {Machine} {Learning}, PMLR, 2019, pp. 3301--3310.
\newline\urlprefix\url{https://proceedings.mlr.press/v97/kaya19a.html}

\bibitem{chen_learning_2020}
X.~Chen, H.~Dai, Y.~Li, X.~Gao, L.~Song, \href{http://arxiv.org/abs/2006.05082}{Learning to {Stop} {While} {Learning} to {Predict}}, arXiv:2006.05082 [cs, stat] (Jun. 2020).
\newblock \href {https://doi.org/10.48550/arXiv.2006.05082} {\path{doi:10.48550/arXiv.2006.05082}}.
\newline\urlprefix\url{http://arxiv.org/abs/2006.05082}

\bibitem{dai_epnet_2020}
X.~Dai, X.~Kong, T.~Guo, \href{https://dl.acm.org/doi/10.1145/3340531.3411973}{{EPNet}: {Learning} to {Exit} with {Flexible} {Multi}-{Branch} {Network}}, in: Proceedings of the 29th {ACM} {International} {Conference} on {Information} \& {Knowledge} {Management}, {CIKM} '20, Association for Computing Machinery, New York, NY, USA, 2020, pp. 235--244.
\newblock \href {https://doi.org/10.1145/3340531.3411973} {\path{doi:10.1145/3340531.3411973}}.
\newline\urlprefix\url{https://dl.acm.org/doi/10.1145/3340531.3411973}

\bibitem{wang_skipnet_2018}
X.~Wang, F.~Yu, Z.-Y. Dou, T.~Darrell, J.~E. Gonzalez, \href{http://arxiv.org/abs/1711.09485}{{SkipNet}: {Learning} {Dynamic} {Routing} in {Convolutional} {Networks}}, arXiv:1711.09485 [cs] (Jul. 2018).
\newline\urlprefix\url{http://arxiv.org/abs/1711.09485}

\bibitem{lin_runtime_2017}
J.~Lin, Y.~Rao, J.~Lu, J.~Zhou, \href{https://papers.nips.cc/paper_files/paper/2017/hash/a51fb975227d6640e4fe47854476d133-Abstract.html}{Runtime {Neural} {Pruning}}, in: Advances in {Neural} {Information} {Processing} {Systems}, Vol.~30, Curran Associates, Inc., 2017.
\newline\urlprefix\url{https://papers.nips.cc/paper_files/paper/2017/hash/a51fb975227d6640e4fe47854476d133-Abstract.html}

\bibitem{fedusSwitchTransformersScaling2022}
W.~Fedus, B.~Zoph, N.~Shazeer, Switch {{Transformers}}: {{Scaling}} to {{Trillion Parameter Models}} with {{Simple}} and {{Efficient Sparsity}} (Jun. 2022).
\newblock \href {http://arxiv.org/abs/2101.03961} {\path{arXiv:2101.03961}}, \href {https://doi.org/10.48550/arXiv.2101.03961} {\path{doi:10.48550/arXiv.2101.03961}}.

\bibitem{hong_panda_2021}
S.~Hong, Y.~Kaya, I.-V. Modoranu, T.~Dumitraş, \href{http://arxiv.org/abs/2010.02432}{A {Panda}? {No}, {It}'s a {Sloth}: {Slowdown} {Attacks} on {Adaptive} {Multi}-{Exit} {Neural} {Network} {Inference}}, arXiv:2010.02432 [cs] (Feb. 2021).
\newblock \href {https://doi.org/10.48550/arXiv.2010.02432} {\path{doi:10.48550/arXiv.2010.02432}}.
\newline\urlprefix\url{http://arxiv.org/abs/2010.02432}

\bibitem{hang_msnet_2023}
R.~Hang, X.~Qian, Q.~Liu, {MSNet}: {Multi}-{Resolution} {Synergistic} {Networks} for {Adaptive} {Inference}, IEEE Transactions on Circuits and Systems for Video Technology 33~(5) (2023) 2009--2018, conference Name: IEEE Transactions on Circuits and Systems for Video Technology.
\newblock \href {https://doi.org/10.1109/TCSVT.2022.3218891} {\path{doi:10.1109/TCSVT.2022.3218891}}.

\bibitem{wangNotAllImages2021}
Y.~Wang, R.~Huang, S.~Song, Z.~Huang, G.~Huang, Not {{All Images}} are {{Worth}} 16x16 {{Words}}: {{Dynamic Transformers}} for {{Efficient Image Recognition}} (Oct. 2021).
\newblock \href {http://arxiv.org/abs/2105.15075} {\path{arXiv:2105.15075}}, \href {https://doi.org/10.48550/arXiv.2105.15075} {\path{doi:10.48550/arXiv.2105.15075}}.

\bibitem{chenCFViTGeneralCoarsetoFine2022}
M.~Chen, M.~Lin, K.~Li, Y.~Shen, Y.~Wu, F.~Chao, R.~Ji, {{CF-ViT}}: {{A General Coarse-to-Fine Method}} for {{Vision Transformer}} (Nov. 2022).
\newblock \href {http://arxiv.org/abs/2203.03821} {\path{arXiv:2203.03821}}, \href {https://doi.org/10.48550/arXiv.2203.03821} {\path{doi:10.48550/arXiv.2203.03821}}.

\bibitem{russakovsky_imagenet_2015}
O.~Russakovsky, J.~Deng, H.~Su, J.~Krause, S.~Satheesh, S.~Ma, Z.~Huang, A.~Karpathy, A.~Khosla, M.~Bernstein, {others}, Imagenet large scale visual recognition challenge, International journal of computer vision 115~(3) (2015) 211--252, publisher: Springer.

\bibitem{riquelmeScalingVisionSparse2021}
C.~Riquelme, J.~Puigcerver, B.~Mustafa, M.~Neumann, R.~Jenatton, A.~S. Pinto, D.~Keysers, N.~Houlsby, Scaling {{Vision}} with {{Sparse Mixture}} of {{Experts}} (Jun. 2021).
\newblock \href {http://arxiv.org/abs/2106.05974} {\path{arXiv:2106.05974}}, \href {https://doi.org/10.48550/arXiv.2106.05974} {\path{doi:10.48550/arXiv.2106.05974}}.

\bibitem{daxbergerMobileVMoEsScaling2023}
E.~Daxberger, F.~Weers, B.~Zhang, T.~Gunter, R.~Pang, M.~Eichner, M.~Emmersberger, Y.~Yang, A.~Toshev, X.~Du, Mobile {{V-MoEs}}: {{Scaling Down Vision Transformers}} via {{Sparse Mixture-of-Experts}} (Sep. 2023).
\newblock \href {http://arxiv.org/abs/2309.04354} {\path{arXiv:2309.04354}}, \href {https://doi.org/10.48550/arXiv.2309.04354} {\path{doi:10.48550/arXiv.2309.04354}}.

\bibitem{he_deep_2016}
K.~He, X.~Zhang, S.~Ren, J.~Sun, Deep residual learning for image recognition, in: Proceedings of the {IEEE} conference on computer vision and pattern recognition, 2016, pp. 770--778.

\bibitem{tan_efficientnet_2019}
M.~Tan, Q.~V. Le, \href{http://proceedings.mlr.press/v97/tan19a.html}{{EfficientNet}: {Rethinking} {Model} {Scaling} for {Convolutional} {Neural} {Networks}}, in: K.~Chaudhuri, R.~Salakhutdinov (Eds.), Proceedings of the 36th {International} {Conference} on {Machine} {Learning}, {ICML} 2019, 9-15 {June} 2019, {Long} {Beach}, {California}, {USA}, Vol.~97 of Proceedings of {Machine} {Learning} {Research}, PMLR, 2019, pp. 6105--6114.
\newline\urlprefix\url{http://proceedings.mlr.press/v97/tan19a.html}

\bibitem{hinton_distilling_2015}
G.~Hinton, O.~Vinyals, J.~Dean, \href{http://arxiv.org/abs/1503.02531}{Distilling the {Knowledge} in a {Neural} {Network}}, arXiv:1503.02531 [cs, stat]ArXiv: 1503.02531 (Mar. 2015).
\newline\urlprefix\url{http://arxiv.org/abs/1503.02531}

\bibitem{lin_fixed_2016}
D.~D. Lin, S.~S. Talathi, V.~S. Annapureddy, \href{http://proceedings.mlr.press/v48/linb16.html}{Fixed {Point} {Quantization} of {Deep} {Convolutional} {Networks}}, in: M.-F. Balcan, K.~Q. Weinberger (Eds.), Proceedings of the 33nd {International} {Conference} on {Machine} {Learning}, {ICML} 2016, {New} {York} {City}, {NY}, {USA}, {June} 19-24, 2016, Vol.~48 of {JMLR} {Workshop} and {Conference} {Proceedings}, JMLR.org, 2016, pp. 2849--2858.
\newline\urlprefix\url{http://proceedings.mlr.press/v48/linb16.html}

\bibitem{han_dynamic_2022}
Y.~Han, G.~Huang, S.~Song, L.~Yang, H.~Wang, Y.~Wang, Dynamic {Neural} {Networks}: {A} {Survey}, IEEE Transactions on Pattern Analysis and Machine Intelligence 44~(11) (2022) 7436--7456, conference Name: IEEE Transactions on Pattern Analysis and Machine Intelligence.
\newblock \href {https://doi.org/10.1109/TPAMI.2021.3117837} {\path{doi:10.1109/TPAMI.2021.3117837}}.

\bibitem{duModelQuantizationHardware2024}
D.~Du, G.~Gong, X.~Chu, Model {{Quantization}} and {{Hardware Acceleration}} for {{Vision Transformers}}: {{A Comprehensive Survey}} (May 2024).
\newblock \href {http://arxiv.org/abs/2405.00314} {\path{arXiv:2405.00314}}.

\bibitem{liu_fastbert_2020}
W.~Liu, P.~Zhou, Z.~Zhao, Z.~Wang, H.~Deng, Q.~Ju, \href{http://arxiv.org/abs/2004.02178}{{FastBERT}: a {Self}-distilling {BERT} with {Adaptive} {Inference} {Time}}, arXiv:2004.02178 [cs] (Apr. 2020).
\newblock \href {https://doi.org/10.48550/arXiv.2004.02178} {\path{doi:10.48550/arXiv.2004.02178}}.
\newline\urlprefix\url{http://arxiv.org/abs/2004.02178}

\bibitem{leontiadis_its_2021}
I.~Leontiadis, S.~Laskaridis, S.~I. Venieris, N.~D. Lane, \href{http://arxiv.org/abs/2102.01393}{It's always personal: {Using} {Early} {Exits} for {Efficient} {On}-{Device} {CNN} {Personalisation}}, in: Proceedings of the 22nd {International} {Workshop} on {Mobile} {Computing} {Systems} and {Applications}, 2021, pp. 15--21, arXiv:2102.01393 [cs].
\newblock \href {https://doi.org/10.1145/3446382.3448359} {\path{doi:10.1145/3446382.3448359}}.
\newline\urlprefix\url{http://arxiv.org/abs/2102.01393}

\bibitem{gardner_artificial_1998}
M.~W. Gardner, S.~R. Dorling, \href{https://www.sciencedirect.com/science/article/pii/S1352231097004470}{Artificial neural networks (the multilayer perceptron)—a review of applications in the atmospheric sciences}, Atmospheric Environment 32~(14) (1998) 2627--2636.
\newblock \href {https://doi.org/10.1016/S1352-2310(97)00447-0} {\path{doi:10.1016/S1352-2310(97)00447-0}}.
\newline\urlprefix\url{https://www.sciencedirect.com/science/article/pii/S1352231097004470}

\bibitem{woo_convnext_2023}
S.~Woo, S.~Debnath, R.~Hu, X.~Chen, Z.~Liu, I.~S. Kweon, S.~Xie, \href{http://arxiv.org/abs/2301.00808}{{ConvNeXt} {V2}: {Co}-designing and {Scaling} {ConvNets} with {Masked} {Autoencoders}}, arXiv:2301.00808 [cs] (Jan. 2023).
\newblock \href {https://doi.org/10.48550/arXiv.2301.00808} {\path{doi:10.48550/arXiv.2301.00808}}.
\newline\urlprefix\url{http://arxiv.org/abs/2301.00808}

\bibitem{ding_davit_2022}
M.~Ding, B.~Xiao, N.~Codella, P.~Luo, J.~Wang, L.~Yuan, {{DaViT}}: {{Dual Attention Vision Transformers}}, in: S.~Avidan, G.~Brostow, M.~Ciss{\'e}, G.~M. Farinella, T.~Hassner (Eds.), Computer {{Vision}} -- {{ECCV}} 2022, Lecture {{Notes}} in {{Computer Science}}, Springer Nature Switzerland, Cham, 2022, pp. 74--92.
\newblock \href {https://doi.org/10.1007/978-3-031-20053-3_5} {\path{doi:10.1007/978-3-031-20053-3_5}}.

\bibitem{team_ResNet_nodate}
P.~Team, \href{https://pytorch.org/hub/pytorch_vision_resnet}{{ResNet} - {PyTorch}}.
\newline\urlprefix\url{https://pytorch.org/hub/pytorch_vision_resnet}

\bibitem{das_recurrent_2023}
S.~Das, A.~Tariq, T.~Santos, S.~S. Kantareddy, I.~Banerjee, Recurrent {{Neural Networks}} ({{RNNs}}): {{Architectures}}, {{Training Tricks}}, and {{Introduction}} to {{Influential Research}}, in: O.~Colliot (Ed.), Machine {{Learning}} for {{Brain Disorders}}, Springer US, New York, NY, 2023, pp. 117--138.
\newblock \href {https://doi.org/10.1007/978-1-0716-3195-9_4} {\path{doi:10.1007/978-1-0716-3195-9_4}}.

\bibitem{vaswani_attention_2017}
A.~Vaswani, N.~Shazeer, N.~Parmar, J.~Uszkoreit, L.~Jones, A.~N. Gomez, L.~Kaiser, I.~Polosukhin, \href{https://papers.nips.cc/paper/2017/hash/3f5ee243547dee91fbd053c1c4a845aa-Abstract.html}{Attention is {All} you {Need}}, in: Advances in {Neural} {Information} {Processing} {Systems}, Vol.~30, Curran Associates, Inc., 2017.
\newline\urlprefix\url{https://papers.nips.cc/paper/2017/hash/3f5ee243547dee91fbd053c1c4a845aa-Abstract.html}

\bibitem{kimDNNPartitioningFramework2024}
H.~Kim, J.~S. Choi, J.~Kim, J.~H. Ko, A {{DNN}} partitioning framework with controlled lossy mechanisms for edge-cloud collaborative intelligence, Future Generation Computer Systems 154 (2024) 426--439.
\newblock \href {https://doi.org/10.1016/j.future.2024.01.006} {\path{doi:10.1016/j.future.2024.01.006}}.

\bibitem{liuAdaptiveDNNInference2023}
G.~Liu, F.~Dai, X.~Xu, X.~Fu, W.~Dou, N.~Kumar, M.~Bilal, An adaptive {{DNN}} inference acceleration framework with end--edge--cloud collaborative computing, Future Generation Computer Systems 140 (2023) 422--435.
\newblock \href {https://doi.org/10.1016/j.future.2022.10.033} {\path{doi:10.1016/j.future.2022.10.033}}.

\end{thebibliography}
\end{document}